\documentclass[conference]{IEEEtran}
\IEEEoverridecommandlockouts
\usepackage{cite}
\usepackage{multirow}
\usepackage{algorithm}
\usepackage{amsmath,amssymb,amsfonts}
\usepackage{algorithmic}
\usepackage{graphicx}
\usepackage{textcomp}
\usepackage{xcolor}
\def\BibTeX{{\rm B\kern-.05em{\sc i\kern-.025em b}\kern-.08em
    T\kern-.1667em\lower.7ex\hbox{E}\kern-.125emX}}
\begin{document}

\title{Semantic-visual Guided Transformer for Few-shot Class-incremental Learning\\
\thanks{$^{*}$Work was done when Sichao Fu was interning at JD Retail POMC. $^{\ddag}$Sichao Fu and Qinmu Peng are the corresponding authors.}
}

\makeatletter
\newcommand{\linebreakand}{%
  \end{@IEEEauthorhalign}
  \hfill\mbox{}\par
  \mbox{}\hfill\begin{@IEEEauthorhalign}
}
\makeatother

\author{
\IEEEauthorblockN{Wenhao Qiu}
\IEEEauthorblockA{\textit{School of Electronic Information and Communications} \\
\textit{Huazhong University of Science and Technology}\\
Wuhan, China \\
qiuwenhao@hust.edu.cn}

\and
\IEEEauthorblockN{Sichao Fu$^{*, \ddag}$}
\IEEEauthorblockA{\textit{School of Electronic Information and Communications} \\
\textit{Huazhong University of Science and Technology}\\
Wuhan, China \\
fusichao$\_$upc@163.com}

\linebreakand % <------------- \and with a line-break
\IEEEauthorblockN{Jingyi Zhang}
\IEEEauthorblockA{\textit{School of Electronic Information and Communications} \\
\textit{Huazhong University of Science and Technology}\\
Wuhan, China \\
zjingyi@hust.edu.cn}
 
\and
\IEEEauthorblockN{Chengxiang Lei}
\IEEEauthorblockA{\textit{School of Electronic Information and Communications} \\
\textit{Huazhong University of Science and Technology}\\
Wuhan, China \\
leichengxiang@hust.edu.cn}
 
\linebreakand % <------------- \and with a line-break
\IEEEauthorblockN{Qinmu Peng$^{\ddag}$}
\IEEEauthorblockA{\textit{School of Electronic Information and Communications} \\
\textit{Huazhong University of Science and Technology}\\
Wuhan, China \\
pengqinmu@hust.edu.cn}
}

\maketitle

\begin{abstract}

Few-shot class-incremental learning (FSCIL) has recently attracted extensive attention in various areas. Existing FSCIL methods highly depend on the robustness of the feature backbone pre-trained on base classes. In recent years, different Transformer variants have obtained significant processes in the feature representation learning of massive fields. Nevertheless, the progress of the Transformer in FSCIL scenarios has not achieved the potential promised in other fields so far. In this paper, we develop a semantic-visual guided Transformer (SV-T) to enhance the feature extracting capacity of the pre-trained feature backbone on incremental classes. Specifically, we first utilize the visual (image) labels provided by the base classes to supervise the optimization of the Transformer. And then, a text encoder is introduced to automatically generate the corresponding semantic (text) labels for each image from the base classes. Finally, the constructed semantic labels are further applied to the Transformer for guiding its hyperparameters updating. Our SV-T can take full advantage of more supervision information from base classes and further enhance the training robustness of the feature backbone. More importantly, our SV-T is an independent method, which can directly apply to the existing FSCIL architectures for acquiring embeddings of various incremental classes. Extensive experiments on three benchmarks, two FSCIL architectures, and two Transformer variants show that our proposed SV-T obtains a significant improvement in comparison to the existing state-of-the-art FSCIL methods.

\end{abstract}

\begin{IEEEkeywords}

Few-shot class-incremental learning, semantic information, visual information, Transformer.

\end{IEEEkeywords}

\section{Introduction}
\label{sec:intro}

Deep learning methods trained by massive supervised information can only give an effective category prediction for seen classes. If we directly finetune the trained model above to simultaneously extract feature embeddings of seen and unseen classes and make label predictions, the model performance of seen categories will have a huge drop \cite{2020maintaining}. In addition, many real-world applications only provide a little supervised information for model re-training. Thus, how to train a deep learning model to continually learn new knowledge from a few unseen classes while avoiding knowledge forgetting of seen classes has become a great challenge. This task is defined as few-shot class-incremental learning (FSCIL) \cite{few-shot, MetaFSCIL, ALICE}.

\begin{figure*}[htbp]
 \centerline{\includegraphics[width=\linewidth]{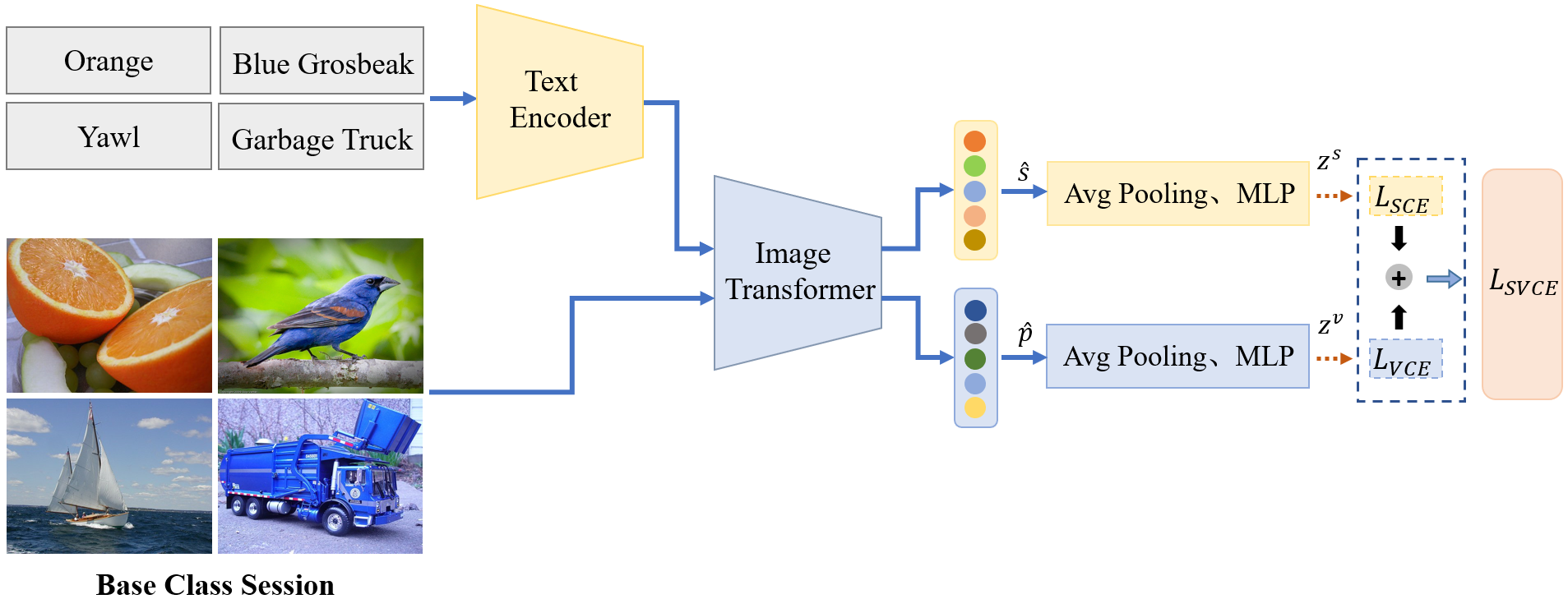}}
 \caption{The basic framework of the proposed semantic-visual guided Transformer (SV-T). The adopted text encoder and image Transformer can be replaced with various variants as needed. It is worth noting that textual semantics are only utilized during the pre-training of the feature backbone.}
 \label{img1}
\end{figure*}

FSCIL has received extensive exploration from researchers at home and abroad since it was first proposed by the TOPIC framework \cite{TOPIC}. Over the past few years, a large number of FSCIL methods have been proposed and also acquired state-of-the-art task performance in many areas. The existing FSCIL methods can be generally divided into two categories. The first category is the pseudo-incremental learning mechanism based on meta-learning \cite{hospedales2021meta}, which divided the provided base class data into massive pseudo-incremental tasks for simulating a real test environment. For example, CEC \cite{CEC} introduced the popular graph attention network to carry out context information propagation between classifiers, named continuously evolving classifier. To enhance the adaption of CEC, authors constructed many pseudo incremental learning tasks via different data augmentation for model optimizations. LIMIT \cite{LIMIT} first synthesized fake FSCIL tasks from the base class data, and then generated a generalized feature space for unseen incremental tasks via multi-phase incremental tasks learning. The second category is to improve the robustness of the model by introducing the self-supervised learning \cite{liu2021self} to construct some auxiliary supervision information. FeSSSS \cite{FeSSSS} first utilized abundant unlabeled data for self-supervised model training. And then, the feature embeddings from the supervised and self-supervised models are fused for the optimization of the FSCIL classifiers. S3C \cite{S3C} proposed a self-supervised stochastic classifier, which regards the rotation angle of the image as an extra self-supervised signal for guiding the optimization of the model.

In recent years, Transformer \cite{han2022survey} has obtained prosperous development and has derived a large number of superior variants in numerous areas, such as Vision Transformer \cite{vit} and Swin Transformer \cite{swinT}. Although massive Transformer variants have been widely adopted in natural language processing (NLP) and computer vision (CV), there is no research attempt to generalize the Transformer that is widely used in the NLP or CV field to FSCIL scenarios so far. If we apply the existing Transformer variants directly, it will cause some serious problems in FSCIL, such as model overfitting and feature blurring. Thus, how to design and train an effective Transformer variant on the base classes to improve its feature extracting capacity on the incremental classes is still a challenging problem. 

To tackle the above challenge, we propose a semantic-visual guided Transformer for few-shot class-incremental learning, named SV-T. Specifically, we utilize a text encoder to automatically learn its semantic labels for given base class data. Through the double guidance of the semantic and visual labels, it can alleviate the model overfitting when generalizing the pre-trained Transformer to the various incremental classes. Finally, the pre-trained Transformer module can be combined with arbitrary exiting FSCIL architectures to obtain the feature embeddings of incremental classes and further make a category prediction.

The contributions of our work are summarized as follows:

\begin{itemize}
\item In this paper, we design an effective Transformer framework pre-trained on base classes for strengthening the feature extracting capacity of FSCIL on the incremental classes. To improve model adaption, our SV-T simultaneously utilizes the semantic and visual labels from the base classes to provide more supervision information for guiding model optimization. 

\item Our SV-T is a simple and flexible module, which can be applied to the arbitrary exiting FSCIL architectures and Transformer variants to extract more effective feature embeddings for any incremental classes.

\item To demonstrate the robustness and scalability of the proposed SV-T in the feature extracting of incremental classes, we conduct extensive experiments on three benchmarks, two FSCIL architectures, and two Transformer variants. The reported results demonstrate that our SV-T outperforms the existing FSCIL methods.
\end{itemize}

\section{Semantic-visual Guided Transformer}
\label{sec:intro}
Recently, several approaches \cite{SEGA} have emerged to exploit semantic knowledge encoded in data to solve few-shot learning tasks, while how to effectively utilize this semantic knowledge to overcome catastrophic forgetting \cite{2017overcoming} of incremental classes in FSCIL has not been explored. In addition, no researchers also have applied the popular Transformer in conjunction with vision and semantics information for FSCIL so far. In this paper, we proposed a novel FSCIL method, which aims to explore the joint use of visual-semantic knowledge and Transformer to improve the adaption of the existing FSCIL models.

\subsection{Problem Formulation}
The main objective of FSCIL is to continuously learn the knowledge of new classes from a few training samples while ensuring the performance of the old categories. In this paper, we utilize $D=\{D^i \}{_{i=0}^N} \in \mathbb{R}^{d_v}$ to denote the training data of different sessions, and then the true labels of the corresponding session $D^i$ is described by $C^i$ ($i \ge 0$). It is worth noting that the labels between different sessions are disjoint, i.e. $\forall i, j$ and $i\neq j$, $C^i\cap C^j = \varnothing$. In our SV-T, we only use the training data in the first (base class) session $D^0$ for Transformer pre-training. And then, the incremental inference is performed to make the pre-trained Transformer adapt to new classes in each subsequent incremental session $D^i$ $(i>0)$. In addition, there only provide a small number of labeled data in subsequent incremental sessions $D^i$ $(i>0)$. For example, in the 10-way 5-shot FSCIL task, each incremental session $D^i$ $(i>0)$ samples 10 new classes from all incremental classes, and each class only has 5 labeled training data. After learning on each incremental session $D^i$, the FSCIL model needs to evaluate the accuracy of the test set overall encountered classes $C^0\cup C^1 \cdots \cup C^i$.

\subsection{Image Transformer and Text Encoder}
As shown in Fig. \ref{img1}, the base class session $D^0$ is used for the Image Transformer pre-train. According to the design of Swin Transformer \cite{swinT}, an input image is first divided into $M$ non-overlapping patches, and the set of these patches is represented as $\{p_i\}{_{i=1}^M}$. Thus, we send these patches to our image Transformer as the start segments. After multi-layers W-MSA and SW-MSA learning in Swin Transformer \cite{swinT}, different image embedding patches $\hat{p} = \{\hat{p}_i\}{_{i=1}^M}$ can be obtained. We further average these outputs, and  use an MLP transformation to obtain the final global visual features. After multi-training, we can get the corresponding visual space $z^v\in \mathbb{R}^{d_v}$.
\begin{equation}
% L_{intra}^{FR}= \frac{1}{2n} \sum_{i=1}^n (L_{intra} (RF_{LGS, i}, RF_{AGS, i}))
\{\hat{p}_i\}{_{i=1}^M} = \emph{Transformer}(\{p_i\}{_{i=1}^M})
\end{equation}
\begin{equation}
z^v = \emph{MLP}(\emph{AvgPool}(\{\hat{p}_i\}{_{i=1}^M}))
\end{equation}

In our FSCIL, the adopted semantic knowledge can come from class labels, attributes, or even knowledge graphs. In this paper, we select the word embeddings of class labels as the semantic knowledge source, i.e. $S=\{s_t \}{_{t=1}^N}$. Where $s_t$ is the label of class t. Then we prompt each of the word embedding label $s_t$ in base class session $D^0$ with the sentence template, e.g., \textit{"A photo of a \{label\}"}. 
% And we use $\mathbb{R}^{d_s}$ to represent the semantic space.

\subsection{Semantic-visual Feature Alignment and Mapping}
Owning to limited training images, the above visual space $V=\{z_i^v \}{_{i=1}^N} \in \mathbb{R}^{d_v}$ is not precise enough, so we utilize semantic knowledge to direct attention to the vision transformer space. We transform the input image and text pairs through our respective encoders and project them into a common embedding space. First, text labels $\{s_t\}{_{t=1}^N}$ are entered into the text encoder to obtain the semantic characteristics of the base class session, and the semantic space $\hat{s}\in \mathbb{R}^{d_s}$ is obtained after the transformer structure. Then we further introduce an MLP to map the feature embeddings in the semantic space $\mathbb{R}^{d_s}$ into the visual space $\mathbb{R}^{d_v}$, that is transformation $f: \mathbb{R}^{d_s} \to \mathbb{R}^{d_v}$. After then, the text features $z^s=f(\hat{s})$ in the common space can be obtained. 

During the training process of base class session $D^0$, assume a batch of ${T_0}$ image-text pairs $\{(\alpha_i^v,\alpha_i^s) \}{_{i=1}^{T_0}}$, where $\alpha_i^v$ and $\alpha_i^s$ are the image and text input of the i-th pairs respectively. We encode each of these as the embedding vectors $z_i^v$ and $z_i^s$ by their respective encoders. The model is then optimized by calculating their cross-entropy losses individually and minimizing both losses simultaneously.
\begin{equation}
L_{VCE}= \frac{1}{T_0} \sum_{i=1}^{T_0} \log\frac{\exp (w_{y_i}^\top z_i^v+b_i)}{\sum_{c\in V} \exp(w_{c}^\top z_c^v+b_c)}
\end{equation}
\begin{equation}
L_{SCE}= \frac{1}{T_0} \sum_{i=1}^{T_0} \log\frac{\exp (w_{y_i}^\top z_i^s+b_i)}{\sum_{c\in S} \exp(w_{c}^\top z_c^s+b_c)}
\end{equation}

Where $T_0$ is the batch size of samples from each batch in the training process, $y_i$ denotes the class corresponding to the $i$-th image-text pair in $T_0$, and $[w_1^\top,...,w_{N}^\top]$ and $[b_1,..., b_{N}]$ are weights and biases in the fully-connected classification layer. Based on the above $ L_{VCE} $ and $L_{SCE}$ losses, the total loss can be defined as follows:
\begin{equation}
L_{SVCE} = L_{VCE} + \lambda \times L_{SCE}
\end{equation}

\begin{algorithm}[t]
\caption{Simplified training procedure of our SV-T}
\label{alg:algorithm}
\begin{algorithmic}[1] %[1] enables line numbers
\STATE \textbf{Input}: Base class session $D_0$ \\
\STATE \textbf{Parameters}: Learning rate $lr_b$, balance parameter $\lambda$, iteration epochs $n$ \\
\STATE \textbf{while} not done \textbf{do}
\STATE  \quad  Get the visual features $z^v$ on $D^0$ via Eq. (1) - Eq. (2)
\STATE  \quad  Generate the semantic labels $\{s_t\}{_{t=1}^N}$ on $D^0$ via sentences template embedding
\STATE  \quad  Get the semantic features $z^s$ on $D^0$ via text encoder and Eq. (1) - Eq. (2)
\STATE  \quad  Calculate the total cross-entropy loss on the visual and semantic features using Eq. (3) - Eq. (5)
\STATE  \quad  Update model hyperparameters via gradient descent
\STATE \textbf{end while} 
\STATE \textbf{Return} the optimal feature backbone trained on $D^0$
\end{algorithmic}
\end{algorithm}

Where $\lambda$ denotes the balance coefficient of $L_{SCE}$ and $L_{VCE}$. This transformation can be understood as a guiding effect on visual features under textual semantics. This approach helps avoid overfitting the network on only a few novel class data. Combining visual and semantic features can achieve strong generalization when training on the base class, which leads to good results when classifying new classes. Algorithm 1 illustrates the simplified training procedure of our proposed SV-T framework.

\begin{table*}[htbp]
    \caption{Experiment comparison with the state-of-the-art FSCIL on the CUB200 dataset. The best results are highlighted.}
    \centering
    \resizebox{\linewidth}{!}{
    \begin{tabular}{c | c c c c c c c c c c c | c c} 
    \hline
    \multirow{2}{*}{Method} & \multicolumn{11}{c|}{Acc. in each session ($\%$)} &  \multirow{2}{*}{Avg.}  & Our relative \\     
    & 0 & 1 & 2 & 3 & 4 & 5 & 6 & 7 & 8 & 9 & 10 &   & improvements \\

    \hline
    CEC (CVPR 2021) \cite{CEC}  & 75.85 & 71.94 & 68.50 & 63.50 & 62.43 & 58.27 & 57.73 & 55.81 & 54.83 & 53.52 & 52.28 & 61.33 & +17.32\\
    MetaFSCIL (CVPR 2022)  \cite{MetaFSCIL}  & 75.90 & 72.41 & 68.78 & 64.78 & 62.96 & 59.99 & 58.30 & 56.85 & 54.78 & 53.82 & 52.64 & 61.93 & +16.72\\
    FeSSSS (CVPR 2022) \cite{FeSSSS} &79.60 & 73.46 & 70.32 & 66.38 & 63.97 & 59.63 & 58.19 & 57.56 & 55.01 & 54.31 & 52.98 & 62.85 & +15.80 \\
    ALICE (ECCV 2022)  \cite{ALICE} &77.40 & 72.70 & 70.60 & 67.20& 65.90 & 63.40 & 62.90 & 61.90 & 60.50 & 60.60 & 60.10 & 65.75 & +12.90\\
    LIMIT (TPAMI 2022)  \cite{LIMIT}  & 75.89 & 73.55 & 71.99 & 68.14 & 67.42 & 63.61 & 62.40 & 61.35 & 59.91 & 58.66 & 57.41 & 65.48 & +13.17\\
    MCNet (TIP 2023) \cite{ji2023memorizing} & 77.57 & 73.96 & 70.47 & 65.81 & 66.16 & 63.81 & 62.09 & 61.82 & 60.41 & 60.09 & 59.08 & 65.57 & +13.08\\
    NC-FSCIL (ICLR 2023) \cite{yang2023neural} & 80.45 & 75.98 & 72.30 & 70.28 & 68.17 & 65.16 & 64.43 & 63.25 & 60.66 & 60.01 & 59.44 & 67.28 & +11.37\\
    SoftNet (ICLR 2023) \cite{kang2022soft} & 78.07 & 74.58 &  71.37 & 67.54 & 65.37 &  62.60 & 61.07 & 59.37 & 57.53 & 57.21 & 56.75 & 64.68 & +13.97 \\
    % SSFE-Net\cite{pan2023ssfe} (WACV 2023) &  76.38 & 72.11 & 68.82 & 64.77 & 63.59 & 60.56 & 59.84 & 58.93 & 57.33 & 56.23 & 54.28 & 62.99 & +15.66 \\
    
    \hline
    
    \textbf{LIMIT+V-Swin-T}  & \textbf{82.59} & \textbf{81.09} & \textbf{79.46} & \textbf{76.68} & \textbf{76.94} & \textbf{75.12} & \textbf{74.59} & \textbf{73.14} & \textbf{73.40} & \textbf{73.17} & \textbf{73.34} & \textbf{76.32} & \textbf{+2.33}  \\
    
    \textbf{LIMIT+SV-Swin-T}  & \textbf{84.19} & \textbf{82.63} & \textbf{81.21} & \textbf{78.97} & \textbf{79.38} & \textbf{77.64} & \textbf{77.55} & \textbf{75.71} & \textbf{75.91} & \textbf{75.77} & \textbf{76.17} & \textbf{78.65} &  \\

    \hline
    \end{tabular}
    }
    \label{table:CUB}
\end{table*}

\begin{table*}[htbp]
    \caption{Experiment comparison with the state-of-the-art FSCIL on the Mini-ImageNet dataset. The best results are highlighted.}
    \centering
    \resizebox{\linewidth}{!}{
    \begin{tabular}{c | c c c c c c c c c | c c} 
    \hline
    \multirow{2}{*}{Method} & \multicolumn{9}{c|}{Acc. in each session ($\%$)} &  \multirow{2}{*}{Avg.}  & Our relative \\     
    & 0 & 1 & 2 & 3 & 4 & 5 & 6 & 7 & 8 &   & improvements \\

    \hline
    CEC (CVPR 2021)  \cite{CEC} & 72.00 & 66.83 & 62.97 & 59.43 & 56.7 & 53.73 & 51.19 & 49.24 & 47.63 & 57.75 & +27.32\\
    MetaFSCIL (CVPR 2022)  \cite{MetaFSCIL} & 72.04 & 67.94 & 63.77 & 60.29 & 57.58 & 55.16 & 52.9 & 50.79 & 49.19 & 58.85 & +26.22\\
    FeSSSS (CVPR 2022)  \cite{FeSSSS} & 81.50 & 77.04 & 72.92 & 69.56 & 67.27 & 64.34 & 62.07 & 60.55 & 58.87 & 68.23 & +16.84\\
    ALICE (ECCV 2022)  \cite{ALICE} &80.60 & 70.60 & 67.40 & 64.50 & 62.50 & 60.00 & 57.80 & 56.80 & 55.70 & 63.99 & +21.08\\
    LIMIT (TPAMI 2022)  \cite{LIMIT} & 72.32 & 68.47 & 64.30 & 60.78 & 57.95 & 55.07 & 52.70 & 50.72 & 49.19 & 59.06 & +26.01\\
    MCNet (TIP 2023) \cite{ji2023memorizing} & 72.33 & 67.70 & 63.50 & 60.34 & 57.59 & 54.70 & 52.13 & 50.41 & 49.08 & 58.64 & +26.43\\
    NC-FSCIL (ICLR 2023) \cite{yang2023neural} & 84.02 & 76.80 & 72.00 & 67.83 & 66.35 & 64.04 & 61.46 & 59.54 & 58.31 & 67.82 &+17.25\\
    SoftNet (ICLR 2023) \cite{kang2022soft} & 79.77 & 75.08 & 70.59 & 66.93 & 64.00 & 61.00 & 57.81 & 55.81 & 54.68 & 65.07 & +20.00 \\
    % SSFE-Net\cite{pan2023ssfe} (WACV 2023) &  72.06 &  66.17 &  62.25 &  59.74 &  56.36 &  53.85 &  51.96 &  49.55 &  47.73 & 57.74 & +27.33 \\
    \hline
    
    \textbf{LIMIT+V-Swin-T} & \textbf{89.17} & \textbf{87.39} & \textbf{84.83} & \textbf{83.41} & \textbf{82.66} & \textbf{81.20} & \textbf{79.81} & \textbf{79.36} & \textbf{79.23} & \textbf{83.01} & \textbf{+2.06} \\
    
    \textbf{LIMIT+SV-Swin-T} & \textbf{90.55} & \textbf{89.20} & \textbf{86.80} & \textbf{85.44} & \textbf{84.78} & \textbf{83.38} & \textbf{81.91} & \textbf{81.90} & \textbf{81.65} & \textbf{85.07}  & \\

    \hline
    \end{tabular}
    }
    \label{table:mini}
\end{table*}

\subsection{Combine with Exiting FSCIL Architectures}
Our proposed SV-T is a simple and independent module, which can be well integrated with the existing FSCIL architectures for incremental class prediction. For example, for FSCIL models with the pseudo-incremental learning mechanism based on meta-learning (such as CEC \cite{CEC} and LIMIT \cite{LIMIT}), our module can be effectively embedded in the process of feature extraction and multi-incremental task training, to train a better classifier that can continue to evolve. Fig. \ref{img2} shows how our module interacts and functions with other FSCIL architectures during the learning process of a new incremental session $D^i$ $(i>0)$. 

\begin{figure}[htbp]
 \centerline{\includegraphics[width=\linewidth]{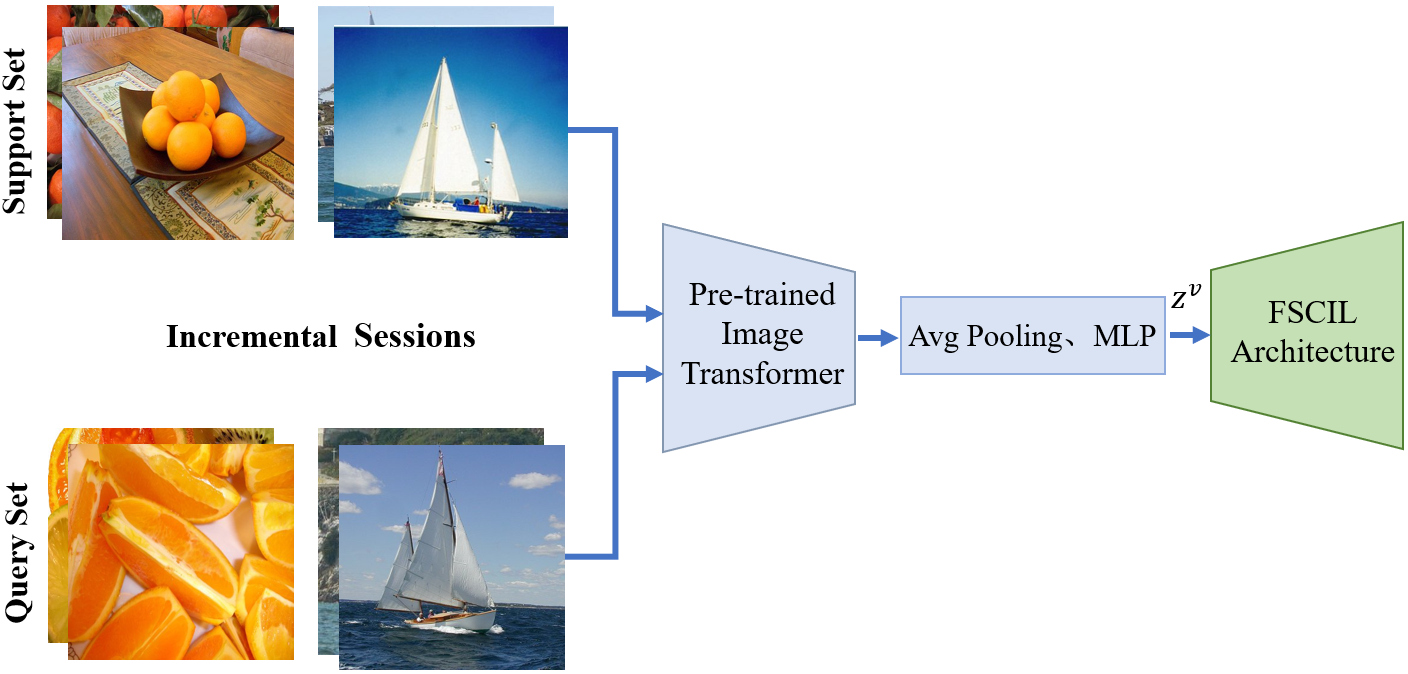}}
 \caption{The incremental inference of the existing FSCIL architectures with the proposed SV-T module on $D^i$ $(i>0)$.}
 \label{img2}
\end{figure}

\section{Experiment}
\label{sec:intro}

\subsection{Datasets}
In this paper, we conduct extensive experiments on three real-world benchmarks including CUB200 \cite{cub}, Mini-ImageNet \cite{mini}, and CIFAR100 \cite{cifar} to validate the effectiveness of the proposed SV-T. 

CUB200 \cite{cub} is a fine-grained image dataset. It is composed of 11788 images with a size of 224×224 RGB collected from 200 bird categories. In our experiments, we follow the same dataset split setting proposed in \cite{TOPIC} and divide each of the 200 classes into 100 base classes and 100 incremental classes. These 100 incremental classes are further divided into 10 sessions, each of which is a 10-way 5-shot task. 

Mini-ImageNet \cite{mini} is a small subset of the ImageNet-2012 dataset. It consists of 100 classes and each class has 600 images, including 500 training images and 100 test images. The image size of Mini-ImageNet is 84-by-84. We first split the 100 classes into 60 base classes and 40 incremental classes. The 40 incremental classes are further divided into 8 incremental sessions, where each session contains five classes and has a 5-way 5-shot incremental task. 

CIFAR100 \cite{cifar} is a popular small-scale image dataset with 60000 images collected from 100 classes. The image is resized to 32 × 32. Each class has 500 images for training and 100 images for testing. Like Mini-ImageNet, we divide these 100 classes into 60 base classes and 40 incremental classes, and CIFAR100 was also adopted in a 5-way 5-shot setting.

\begin{table*}[htbp]
    \caption{Experiment comparison with the state-of-the-art FSCIL on the CIFAR100 dataset. The best results are highlighted.}
    \centering
    \resizebox{\linewidth}{!}{
    \begin{tabular}{c | c c c c c c c c c | c c} 
    \hline
    \multirow{2}{*}{Method} & \multicolumn{9}{c|}{Acc. in each session ($\%$)} &  \multirow{2}{*}{Avg.}  & Our relative \\     
    & 0 & 1 & 2 & 3 & 4 & 5 & 6 & 7 & 8 &   & improvements \\

    \hline
    CEC (CVPR 2021)  \cite{CEC} &73.07 & 68.88 & 65.26 & 61.19 & 58.09 & 55.57 & 53.22 & 51.34 & 49.14 & 59.53 & +17.31\\
    MetaFSCIL (CVPR 2022)  \cite{MetaFSCIL} & 74.50 & 70.10 & 66.84 & 62.77 & 59.48 & 56.52 & 54.36 & 52.56 & 49.97 & 60.79 & +16.05\\
    FeSSSS (CVPR2022)  \cite{FeSSSS} & 75.35 & 70.81 & 66.7 & 62.73 & 59.62 & 56.45 & 54.33 & 52.10 & 50.23 & 60.92 & +15.92\\
    ALICE (ECCV 2022)  \cite{ALICE} &79.00 & 70.50 & 67.10 & 63.40 & 61.20 & 59.20 & 58.10 & 56.30 & 54.10 & 63.21 & +13.63\\
    LIMIT (TPAMI 2022)  \cite{LIMIT} & 73.81 & 72.09 & 67.87 & 63.89 & 60.70 & 57.77 & 55.67 & 53.52 & 51.23 & 61.84 & +15.00\\
    MCNet (TIP 2023) \cite{ji2023memorizing} & 73.30 & 69.34 & 65.72 & 61.70 & 58.75 & 56.44 & 54.59 & 53.01 & 50.72 & 60.40 & +16.44\\
    SoftNet (ICLR 2023) \cite{kang2022soft} & 80.33 & 76.23 & 72.19 & 67.83 & 64.64 & 61.39 &  59.32 & 57.37 & 54.94 & 66.03 & +10.81\\
    NC-FSCIL (ICLR 2023) \cite{yang2023neural} & 82.52 & 76.82 & 73.34 & 69.68 & 66.19 & 62.85 & 60.96 & 59.02 & 56.11 & 67.50 & +9.34\\
    \hline
    
    \textbf{LIMIT+V-Swin-T}   & \textbf{82.07} & \textbf{78.49} & \textbf{75.90} & \textbf{73.27} & \textbf{72.36} & \textbf{71.20} & \textbf{70.60} & \textbf{69.39} & \textbf{67.69} & \textbf{73.44} & \textbf{+3.40} \\
    \textbf{LIMIT+SV-Swin-T}   & \textbf{86.77} & \textbf{82.82} & \textbf{80.36} & \textbf{77.20} & \textbf{76.06} & \textbf{74.00} & \textbf{72.92} & \textbf{71.68} & \textbf{69.75} & \textbf{76.84} &  \\

    \hline
    \end{tabular}
    }
    \label{table:cifar}
\end{table*}

\begin{figure*}[htbp]
\begin{minipage}[b]{0.245\linewidth}
  \centering
  \includegraphics[width=\linewidth]{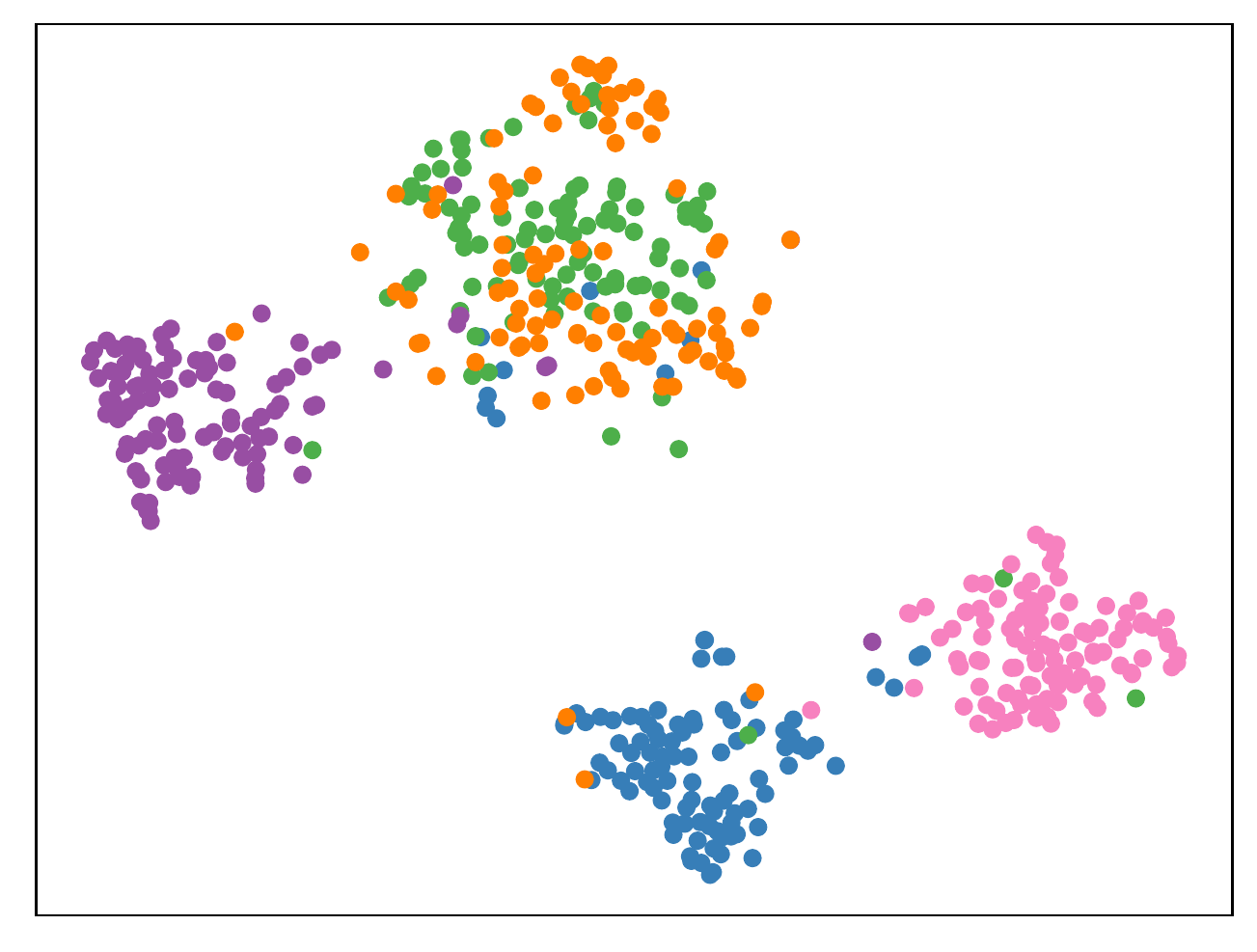}
  \centerline{(a) CEC}\medskip
\end{minipage}
\hfill
\begin{minipage}[b]{0.245\linewidth}
  \centering
  \includegraphics[width=\linewidth]{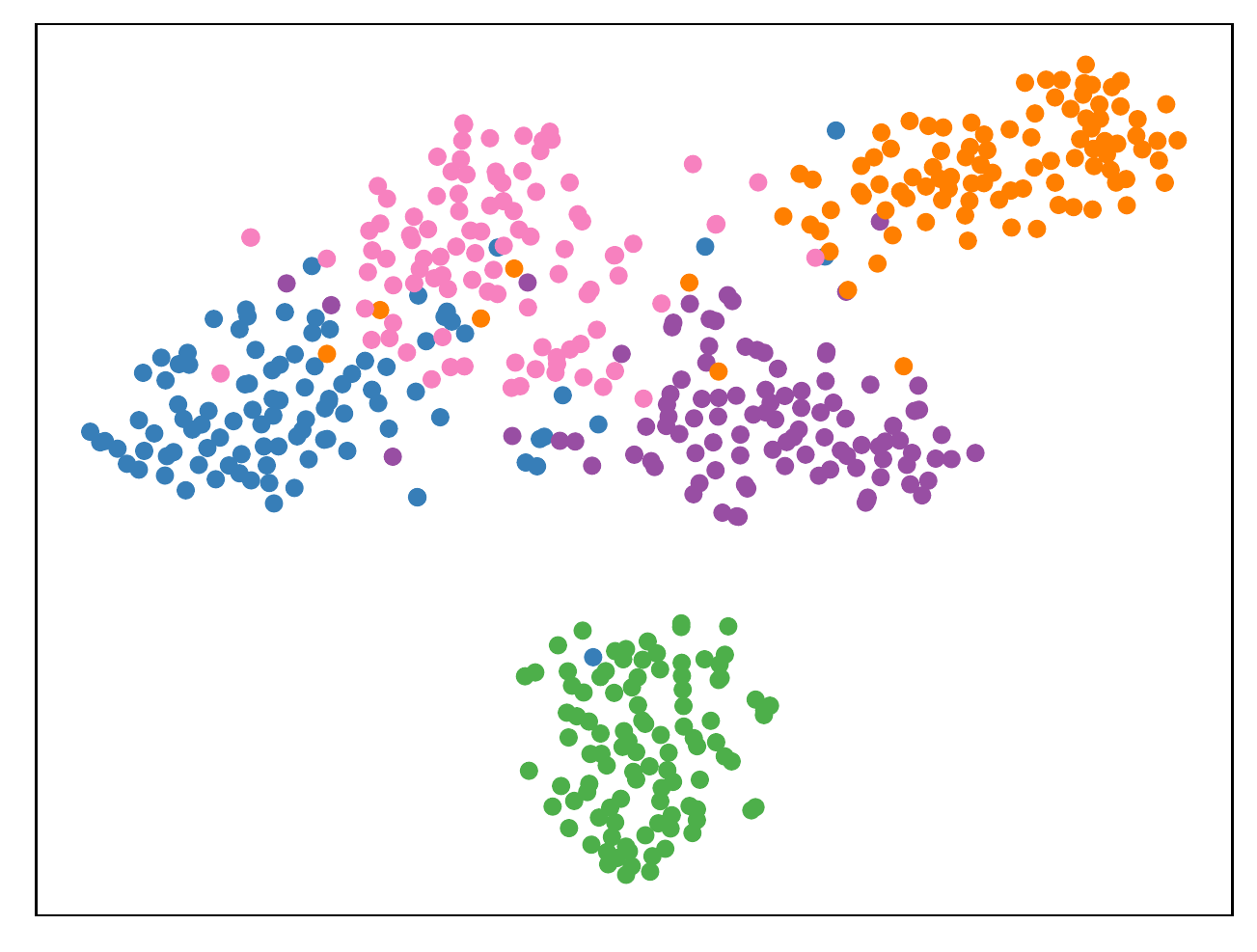}
  \centerline{(b) LIMIT}\medskip
\end{minipage}
\begin{minipage}[b]{0.245\linewidth}
  \centering
  \includegraphics[width=\linewidth]{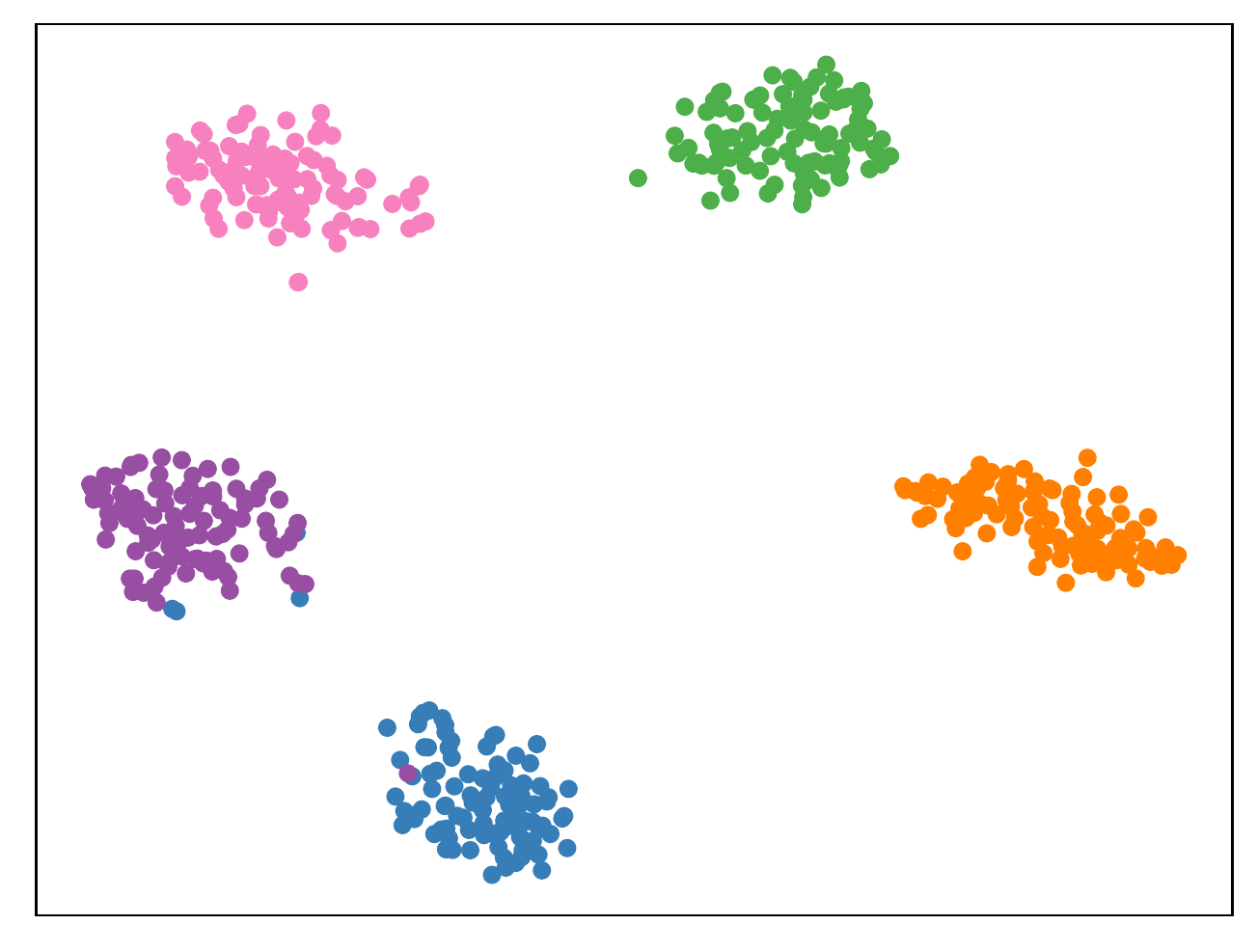}
  \centerline{(c) LIMIT+V-Swin-T}\medskip
\end{minipage}
\begin{minipage}[b]{0.245\linewidth}
  \centering
  \includegraphics[width=\linewidth]{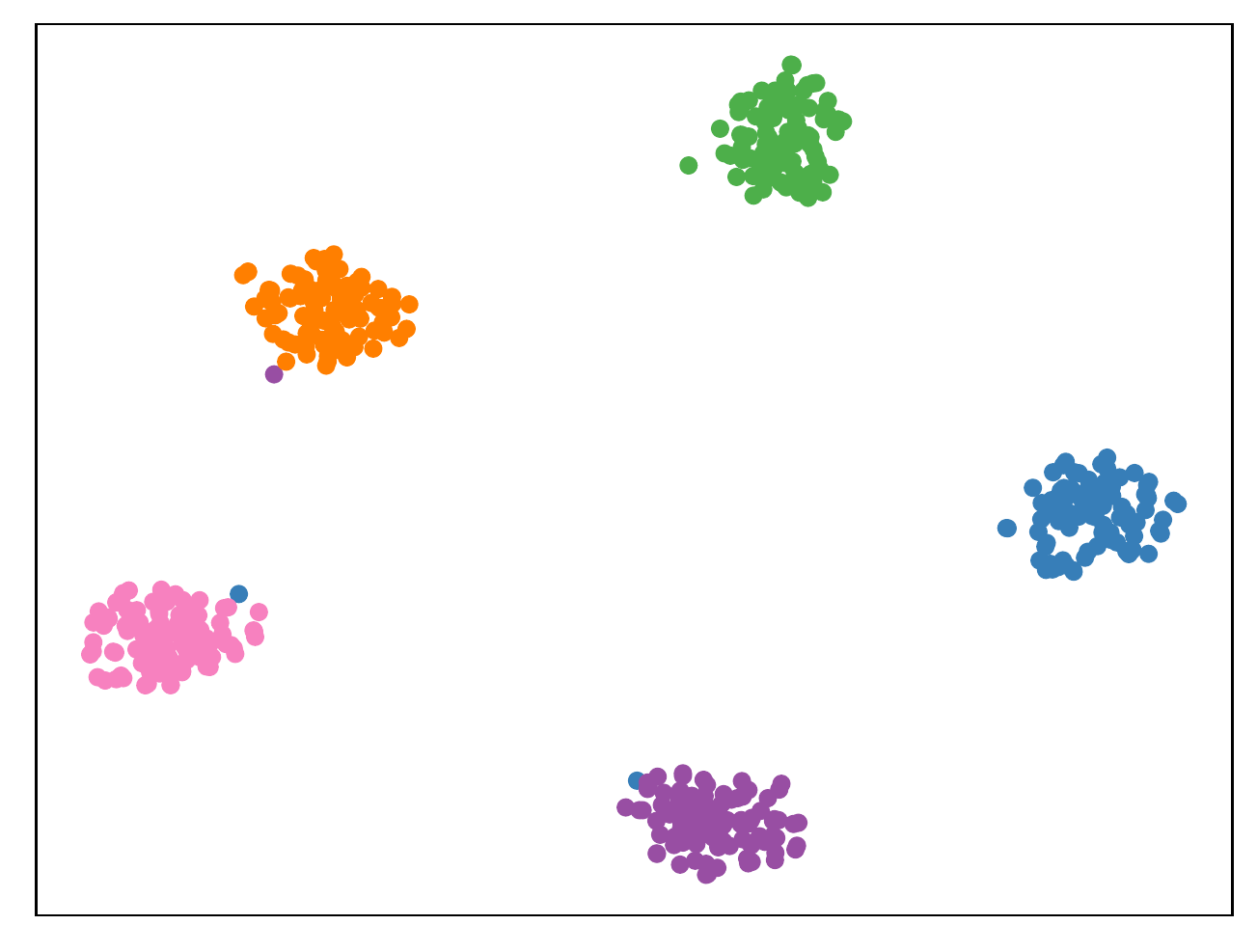}
  \centerline{(d) LIMIT+SV-Swin-T}\medskip
\end{minipage}
\caption{t-SNE feature embeddings of the 8-th session on the CIFAR100 dataset. Each of the five colors represents a different class: \emph{blue} = whale, \emph{green} = willow tree, \emph{purple} = wolf, \emph{orange} = woman and \emph{pink} = worm.}
\label{img3}
\end{figure*}

\subsection{Experimental Setup}
Our SV-T is a simple and flexible module, which can be well combined with any existing FSCIL architecture. In this paper, we use the recent state-of-the-art FSCIL architecture-LIMIT \cite{LIMIT} as our baseline model. Specifically, we regard the popular Swin Transformer (Swin-T) \cite{swinT} as the basic feature module (\textbf{Note: our SV-T will be re-named SV-Swin-T}), and then regard the Swin-Tiny parameters pre-trained by the ImageNet-22K dataset as initializing parameters of the image Transformer. Owning to the excellent performance of the CLIP \cite{clip} model in multi-modal domains and zero-shot directions, the advanced CLIP framework is introduced as our text encoder module. In our SV-T, CLIP parameters pre-trained on ViT-B/32 \cite{vit} is introduced to initialize the text encoder.

During model training, we only utilize the visual and semantic labels of the base classes for hyperparameter adjustment. The SGD with the learning rate $lr_b$ of 0.01 and momentum is introduced for model training. The maximum iteration epochs of the proposed SV-T are set to 500. We trained each model another 100 epochs for the learning rate of 0.0002, with a decay of 0.5 every 100 iterations. In our SV-T, we follow the same preprocessing strategies described by LIMIT \cite{LIMIT}, such as random scaling, random cropping, and random horizontal flipping.

\subsection{Experimental Results and Analysis}
\subsubsection{Comparison with State-of-the-art Methods}
In Tables \uppercase\expandafter{\romannumeral1}-\uppercase\expandafter{\romannumeral3}, we show the classification performance of our proposed SV-T and recent state-of-the-art FSCIL methods on the CIFAR100, Mini-Imagenet, and CUB200 datasets. In our experiments, we introduce the widely-used Top-1 accuracy of each session and average accuracy rate $\textbf{Avg.}=\frac{1}{N} \sum_{i} A_{i} $ of all sessions to estimate model performance. $A_i$ denotes the Top-1 accuracy in the i-th session. From these results, we can observe that our proposed LIMIT+SV-Swin-T obtains the best classification performance on all benchmarks. Specifically, compared to the baseline LIMIT (Avg.), our LIMIT+SV-Swin-T achieve gains of 26.01$\%$ on the Mini-Imagenet dataset and 15$\%$ on the CIFAR100 dataset, respectively. More importantly, our proposed LIMIT+SV-Swin-T not only achieves a large improvement in Avg. but also has a significant improvement in the accuracy of adapting to incremental sessions, which directly demonstrates the robust feature extracting capacity of our SV-T on incremental classes. 

\subsubsection{Ablation Studies}
In this part, we conduct extensive experiments to validate the effect of each component (visual guided Transformer (V-T) and semantic-visual guided Transformer (SV-T)) for FSCIL. As shown in Tables \uppercase\expandafter{\romannumeral1}-\uppercase\expandafter{\romannumeral3}, we can find that LIMIT+V-Swin-T with the guidance of visual labels all outperform the recent FSCIL in Top-1 and Avg., which directly demonstrates the superiority of the existing Transformer on incremental classes of FSCIL. In addition, our LIMIT+SV-Swin-T further improves the classification performance of LIMIT+V-Swin-T under the double guidance of semantic and visual labels. For example, LIMIT+SV-Swin-T outperforms LIMIT+V-Swin-T by 2.33$\%$ on the CUB200 dataset, 3.4$\%$ on the CIFAR100 dataset, respectively. These results show that reasonably utilizing more supervision information from the data itself can further improve the robustness of the pre-trained feature backbone. 

\subsubsection{t-SNE Visualization}
To further demonstrate that our V-T and SV-T can better learn the feature knowledge of incremental classes, we introduce the widely-used t-SNE tool to visualize the feature distribution map in 2-D space. For t-SNE, sample features belonging to the same category are expected to cluster together. Fig. \ref{img3} reveals that our proposed LIMIT+V-T and LIMIT+SV-T all better separate different classes and also make the same class clustered together more compactly in comparison to classic CEC \cite{CEC} and LIMIT \cite{LIMIT}. This visual rendering further confirms the effectiveness of our approach.

\begin{figure}[htbp]
\begin{minipage}[b]{0.49\linewidth}
  \centering
  \includegraphics[width=\linewidth]{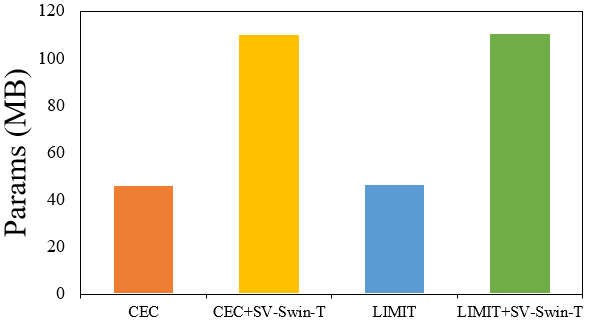}
  \centerline{(a) Params}\medskip
\end{minipage}
\hfill
\begin{minipage}[b]{0.49\linewidth}
  \centering
  \includegraphics[width=\linewidth]{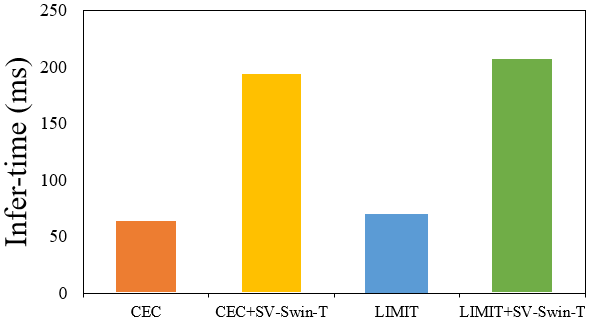}
  \centerline{(b) Infer-time}\medskip
\end{minipage}
\caption{Experiment comparison on the CUB200 dataset. MB and ms denote Megabyte and Millisecond. The infer-time is measured on NVIDIA GTX 1080Ti with batch size 16 for every 100 images in the test set.}
\label{img4}
\end{figure}

\subsubsection{Computational Complexity}

In Fig. \ref{img4}, we conduct extensive experiments to analyze the model parameters (Params) and the infer-time of the proposed LIMIT+SV-Swin-T, CEC+SV-Swin-T, and two representative FSCIL methods. From these results, we can see that this kind of the consumption on the GPU is acceptable, although our proposed method needs more Params, and infer-time in Fig. \ref{img4}. Thus, our proposed method may be a feasible way in reality.

\section{CONCLUSION}
In this paper, we propose a semantic-visual guided Transformer for FSCIL. Our proposed SV-T framework obtains a significant performance improvement in overcoming catastrophic forgetting of incremental classes. Extensive experiments on three well-known benchmarks demonstrate the superiority of the proposed method. What is more important, our SV-T can be well-compatible with arbitrary FSCIL architectures to improve their classification performance. 

\section*{Acknowledgment}

This work was supported in part by the National Key Research and Development Program of China under Grant 2022YFF0712300, 2022YFC3301004, 2022YFC3301003 and 2022YFC3301704, in part by the Fundamental Research Funds for the Central Universities (HUST) under Grant 2022JYCXJJ034 and YCJJ202204016, in part by the Open Research Fund from Shandong Provincial Key Laboratory of Computer Network under Grant SKLCN-2021-02.

\section{Appendix}

\subsection{Model Scalability}

In the supplementary section, we provide more experiment results to reflect the completeness of the experiments we have done. In Tables \uppercase\expandafter{\romannumeral4}-\uppercase\expandafter{\romannumeral5}, we list the experiment results from a large number of FSCIL models proposed in recent years. In this subsection, the widely-used Vision Transformer (ViT) and Swin Transformer (Swin-T) are first introduced to regard as different Transformer variants in Table \uppercase\expandafter{\romannumeral4} (\textbf{Note: our SV-T will be re-named SV-ViT and SV-Swin-T in turn}). In Table \uppercase\expandafter{\romannumeral5}, we further introduce the classic CEC \cite{CEC} and LIMIT \cite{LIMIT} model as our baselines. In this paper, we uniformly adopt the widely-used Top-1 accuracy of each session and the average accuracy of all sessions as the measurement index. In addition, we also conduct extensive experiments to validate the effect of each component (visual guided Transformer (V-T) and semantic-visual guided Transformer (SV-T)) for different Transformer variants and FSCIL architectures. The experiment results in Table \uppercase\expandafter{\romannumeral4} show that our proposed SV-T method can be compatible with different Transformer variants and can obtain a novel state-of-the-art FSCIL classification performance in all sessions. The experiment results in Table \uppercase\expandafter{\romannumeral5} further demonstrate that the proposed SV-T is a simple and flexible module, which can be effectively applied to the arbitrarily FSCIL architectures for avoiding knowledge forgetting and overfitting problem of the pre-trained feature backbone on the incremental classes. As shown in Tables \uppercase\expandafter{\romannumeral4}-\uppercase\expandafter{\romannumeral5}, we can observe that our SV-T (LIMIT+SV-ViT, LIMIT+SV-Swin-T, CEC+SV-Swin-T) with the double guidance of semantic and visual labels can better improve the robustness of the pre-train feature backbone on the incremental classes in comparison to simply relying on semantic or visual labels from data itself (LIMIT+V-ViT, LIMIT+V-Swin-T, CEC+V-Swin-T).

\subsection{t-SNE Visualization}

In this part, we further utilize the widely-used t-SNE method to visualize the feature embedding distribution of the $i$-th FSCIL session in two-dimension space for validating whether our approach can effectively be combined with any FSCIL architectures and different Transformer variants (\textbf{Note: our SV-T will be re-named SV-ViT and SV-Swin-T in turn}). In the t-SNE tool, feature embeddings belonging to the same class are expected to be clustered together. Fig. \ref{img} shows that our proposed SV-T model can more accurately separate different classes than either classical CEC or LIMIT methods. In addition, no matter whether using ViT (Swin-T) as our Transformer variants or CEC (LIMIT) as our FSCIL basslines, the proposed SV-T all achieve the best classification performance. In addition, FSCIL with the Swin-T module has slightly better visualization results than FSCIL with the ViT module due to its superior performance. These results further demonstrate the wide applicability and robustness of the proposed SV-T method.

\begin{table*}[bp]
    \caption{Experiment comparison with the state-of-the-art FSCIL on the CUB200 dataset. The best results are highlighted.}
    \centering
    \resizebox{\linewidth}{!}{
    \begin{tabular}{c | c c c c c c c c c c c | c c} 
    \hline
    \multirow{2}{*}{Method} & \multicolumn{11}{c|}{Acc. in each session ($\%$)} &  \multirow{2}{*}{Avg.}  & Our relative \\     
    & 0 & 1 & 2 & 3 & 4 & 5 & 6 & 7 & 8 & 9 & 10 &   & improvements \\

    \hline
    Ft-CNN  & 68.68 & 43.70 & 25.05 & 17.72 & 18.08 & 16.95 & 15.10 & 10.60 & 8.93 & 8.93 & 8.47 & 22.02 & +56.63\\
    iCaRL (CVPR 2017) \cite{iCaRL(CVPR2017)}  & 68.68 & 52.65 & 48.61 & 44.16 & 36.62 & 29.52 & 27.83 & 26.26 & 24.01 & 23.89 & 21.16 & 36.67 &  +41.98\\
    EEIL (ECCV 2018) \cite{EEil(ECCV2018)}  & 68.68 & 53.63 & 47.91 & 44.20 & 36.30 & 27.46 & 25.93 & 24.70 & 23.95 & 24.13 & 22.11 & 36.27 &  +42.38\\
    NCM (CVPR 2019)  \cite{NCM(CVPR2019)}  & 68.68 & 57.12 & 44.21 & 28.78 & 26.71 & 25.66 & 24.62 & 21.52 & 20.12 & 20.06 & 19.87 & 32.49 & +46.16\\
    TOPIC (CVPR 2020) \cite{TOPIC(CVPR2020)}  & 68.68 & 62.49 & 54.81 & 49.99 & 45.25 & 41.40 & 38.35 & 35.36 & 32.22 & 28.31 & 26.28 & 43.92 &  +34.73\\
    Decoupled-DeepEMD (CVPR 2020) \cite{DeepDEM(CVPR2020)} & 75.35 & 70.69 & 66.68 & 62.34 & 59.76 & 56.54 & 54.61 & 52.52 & 50.73 & 49.20 & 47.60 & 58.73 & +19.92\\
    SS-iCaRL (ICIP 2021) \cite{SS-icarl(ICIP2021)} & 69.89 & 61.24 & 55.81 & 50.99 & 48.18 & 46.91 & 43.99 & 39.78 & 37.50 & 34.54 & 31.33 & 47.28 &  +31.37\\
    SS-NCM (ICIP 2021) \cite{SS-icarl(ICIP2021)} & 69.89 & 61.91 & 55.51 & 51.71 & 49.68 & 46.11 & 42.19 & 39.03 & 37.96 & 34.05 & 32.65 & 47.33 &  +31.32\\
    SS-NCM-CNN (ICIP 2021) \cite{SS-icarl(ICIP2021)} & 69.89 & 64.87 & 59.82 & 55.14 & 52.48 & 49.60 & 47.87 & 45.10 & 40.47 & 38.10 & 35.25 & 50.78 &  +27.87\\
    Cheraghian et.al (ICCV 2021)  \cite{Cheraghian_2021_ICCV}  & 68.78 & 59.37 & 59.32 & 54.96 & 52.58 & 49.81 & 48.09 & 46.32 & 44.33 & 43.43 & 43.23 & 51.84 & +33.23\\
    SAKD (CVPR 2021) \cite{Cheraghian_2021_CVPR}  & 68.23 & 60.45 & 55.70 & 50.45 & 45.72 & 42.90 & 40.89 & 38.77 & 36.51 & 34.87 & 32.96 & 46.13 & +38.94 \\
    SPPR (CVPR 2021) \cite{SPPR} & 68.68 & 61.85 & 57.43 & 52.68 & 50.19 & 46.88 & 44.65 & 43.07 & 40.17 & 39.63 & 37.33 & 49.32 &  +29.33\\
    CEC (CVPR 2021)  \cite{CEC}  & 75.85 & 71.94 & 68.50 & 63.50 & 62.43 & 58.27 & 57.73 & 55.81 & 54.83 & 53.52 & 52.28 & 61.33 & +17.32 \\
    ERL (AAAI 2021) \cite{ERL(AAAI2021)} & 73.52 & 70.12 & 65.12 & 62.01 & 58.56 & 57.99 & 56.77 & 56.52 & 55.01 & 53.68 & 50.01 & 59.93 &  +18.72\\
    ERL++ (AAAI 2021) \cite{ERL(AAAI2021)} & 73.52 & 71.09 & 66.13 & 63.25 & 59.49 & 59.89 & 58.64 & 57.72 & 56.15 & 54.75 & 52.28 & 61.18 &  +17.47\\
    FSLL+SS (AAAI 2021)  \cite{FSLL+SS} & 75.63 & 71.81 & 68.16 & 64.32 & 62.61 & 60.10 & 58.82 & 58.70 & 56.45 & 56.41 & 55.82 & 62.62 & +16.03\\
    F2M (NIPS 2021)  \cite{F2M} & 77.13 & 73.92 & 70.27 & 66.37 & 64.34 & 61.69 & 60.52 & 59.38 & 57.15 & 56.94 & 55.89 & 63.96 & +14.69\\
    IDLVQ-C (ICLR 2021)  \cite{IDLVQ-C} & 77.37 & 74.72 & 70.28 & 67.13 & 65.34 & 63.52 & 62.10 & 61.54 & 59.04 & 58.68 & 57.81 & 65.23 & +13.42\\
    MetaFSCIL (CVPR 2022)  \cite{MetaFSCIL}  & 75.90 & 72.41 & 68.78 & 64.78 & 62.96 & 59.99 & 58.30 & 56.85 & 54.78 & 53.82 & 52.64 & 61.93 & +16.72\\
    FeSSSS (CVPR 2022)  \cite{FeSSSS} &79.60 & 73.46 & 70.32 & 66.38 & 63.97 & 59.63 & 58.19 & 57.56 & 55.01 & 54.31 & 52.98 & 62.85 & +15.80\\
    FACT (CVPR 2022)  \cite{FACT}  & 75.90 & 73.23 & 70.84 & 66.13 & 65.56 & 62.15 & 61.74 & 59.83 & 58.41 & 57.89 & 56.94 & 64.42 & +14.23\\
    ALICE (ECCV 2022)  \cite{ALICE} &77.40 & 72.70 & 70.60 & 67.20 & 65.90 & 63.40 & 62.90 & 61.90 & 60.50 & 60.60 & 60.10 & 65.75 & +12.90\\
    S3C (ECCV 2022)  \cite{S3C} & 80.62 & 77.55 & 73.19 & 68.54 & 68.05 & 64.33 & 63.58 & 62.07 & 60.61 & 59.79 & 58.95 & 67.03 & +11.62\\
    Liu et al. (ECCV 2022) \cite{liu2022few} & 75.90 & 72.14 & 68.64 & 63.76 & 62.58 & 59.11 & 57.82 & 55.89 & 54.92 & 53.58 & 52.39 & 61.52 & +17.13 \\
    SvF (TPAMI 2022) \cite{zhao2021mgsvf} & 72.29 & 70.53 & 67.00 &  64.92 &  62.67 & 61.89 & 59.63 & 59.15 & 57.73 & 55.92 &  54.33 & 62.37 & +16.28\\
    LIMIT (TPAMI 2022)  \cite{LIMIT}  & 75.89 & 73.55 & 71.99 & 68.14 & 67.42 & 63.61 & 62.40 & 61.35 & 59.91 & 58.66 & 57.41 & 65.48 & +13.17\\
    MCNet (TIP 2023) \cite{ji2023memorizing} & 77.57 & 73.96 & 70.47 & 65.81 & 66.16 & 63.81 & 62.09 & 61.82 & 60.41 & 60.09 & 59.08 & 65.57 & +13.08\\
    NC-FSCIL (ICLR 2023) \cite{yang2023neural} & 80.45 & 75.98 & 72.30 & 70.28 & 68.17 & 65.16 & 64.43 & 63.25 & 60.66 & 60.01 & 59.44 & 67.28 & +11.37\\
    SoftNet (ICLR 2023) \cite{kang2022soft} & 78.07 & 74.58 &  71.37 & 67.54 & 65.37 &  62.60 & 61.07 & 59.37 & 57.53 & 57.21 & 56.75 & 64.68 & +13.97 \\
    \hline

    \textbf{LIMIT+V-ViT}  & \textbf{80.30} & \textbf{78.41} & \textbf{77.32} & \textbf{74.77} & \textbf{74.96} & \textbf{73.16} & \textbf{72.78} & \textbf{71.13} & \textbf{71.06} & \textbf{70.92} & \textbf{71.02} & \textbf{74.17} & \textbf{+4.48} \\
    \textbf{LIMIT+SV-ViT}  & \textbf{82.20} & \textbf{80.31} & \textbf{79.18} & \textbf{76.60} & \textbf{76.82} & \textbf{75.15} & \textbf{74.78} & \textbf{73.99} & \textbf{73.83} & \textbf{73.43} & \textbf{73.52} & \textbf{76.35} & \textbf{+2.30} \\

    \hline
    \textbf{LIMIT+V-Swin-T}  & \textbf{82.59} & \textbf{81.09} & \textbf{79.46} & \textbf{76.68} & \textbf{76.94} & \textbf{75.12} & \textbf{74.59} & \textbf{73.14} & \textbf{73.40} & \textbf{73.17} & \textbf{73.34} & \textbf{76.32} & \textbf{+2.33}  \\
    
    \textbf{LIMIT+SV-Swin-T} & \textbf{84.19} & \textbf{82.63} & \textbf{81.21} & \textbf{78.97} & \textbf{79.38} & \textbf{77.64} & \textbf{77.55} & \textbf{75.71} & \textbf{75.91} & \textbf{75.77} & \textbf{76.17} & \textbf{78.65} &  \\

    \hline
    \end{tabular}
    }
    \label{table:compare FP}
\end{table*}

\begin{table*}[bp]
    \caption{Experiment comparison with the state-of-the-art FSCIL on the Mini-ImageNet dataset. The best results are highlighted.}
    \centering
    \resizebox{\linewidth}{!}{
    \begin{tabular}{c | c c c c c c c c c | c c} 
    \hline
    \multirow{2}{*}{Method} & \multicolumn{9}{c|}{Acc. in each session ($\%$)} &  \multirow{2}{*}{Avg.}  & Our relative \\     
    & 0 & 1 & 2 & 3 & 4 & 5 & 6 & 7 & 8  &   & improvements \\

    \hline
    Ft-CNN   & 61.31 & 27.22 & 16.37 & 6.08 & 2.54 & 1.56 & 1.93 & 2.60 & 1.40 & 13.44 & +71.63 \\
    iCaRL (CVPR 2017) \cite{iCaRL(CVPR2017)} & 61.31 & 46.32 & 42.94 & 37.63 & 30.49 & 24.00 & 20.89 & 18.80 & 17.21 & 33.28 & +51.79\\
    EEIL (ECCV 2018) \cite{EEil(ECCV2018)} & 61.31 & 46.58 & 44.00 & 37.29 & 33.14 & 27.12 & 24.10 & 21.57 & 19.58 & 34.96 & +50.11\\
    NCM (CVPR 2019) \cite{NCM(CVPR2019)} & 61.31 & 47.80 & 39.31 & 31.91 & 25.68 & 21.35 & 18.67 & 17.24 & 14.17 & 30.82 & +54.25\\
    TOPIC (CVPR 2020) \cite{TOPIC(CVPR2020)} & 61.31 & 50.09 & 45.17 & 41.16 & 37.48 & 35.52 & 32.19 & 29.46 & 24.42 & 39.64 & +45.43\\
    Decoupled-DeepEMD (CVPR 2020) \cite{DeepDEM(CVPR2020)} & 69.77 & 64.59 & 60.21 & 56.63 & 53.16 & 50.13 & 47.49 & 45.42 & 43.41 & 54.53 & +30.54\\
    SS-NCM-CNN (ICIP 2021) \cite{SS-icarl(ICIP2021)}  & 62.88 & 60.66 & 57.55 & 52.66 & 50.44 & 48.44 & 45.11 & 41.55 & 40.88 & 51.13 & +33.94\\
    Cheraghian et.al (ICCV 2021) \cite{Cheraghian_2021_ICCV} & 61.40 & 59.80 & 54.20 & 51.69 & 49.45 & 48.00 & 45.20 & 43.80 & 42.10 & 50.63 & +34.44 \\
    SAKD (CVPR 2021)  \cite{Cheraghian_2021_CVPR}  &62.00 & 58.00 & 52.00 & 49.00 & 48.00 & 45.00 & 42.00 & 40.00 & 39.00 & 48.33 & +36.74\\
    SPPR (CVPR 2021)   \cite{SPPR} & 80.00 & 74.00 & 68.66 & 64.33 & 61.00 & 57.33 & 54.66 & 51.66 & 49.00 & 62.29 & +22.78\\
    CEC (CVPR 2021)  \cite{CEC} & 72.00 & 66.83 & 62.97 & 59.43 & 56.70 & 53.73 & 51.19 & 49.24 & 47.63 & 57.75 & +27.32\\
    ERL (AAAI 2021) \cite{ERL(AAAI2021)} & 61.67 & 56.19 & 54.70 & 51.19 & 47.61 & 45.23 & 44.00 & 40.95 & 39.80 & 49.03 & +36.04\\
    ERL++ (AAAI 2021) \cite{ERL(AAAI2021)} & 61.67 & 57.61 & 54.76 & 51.67 & 48.57 & 46.42 & 44.04 & 42.85 & 40.71 & 49.81 & +35.26\\
    IDLVQ-C (ICLR 2021)  \cite{IDLVQ-C} & 64.77 & 59.87 & 55.93 & 52.62 & 49.88 & 47.55 & 44.83 & 43.14 & 41.84 & 51.16 & +33.91\\
    LCwoF (ICCV 2021)  \cite{ICwoF} & 64.45 & 59.88 & 56.10 & 52.75 & 50.20 & 47.71 & 44.97 & 43.74 & 42.84 & 51.40 & +33.67\\
    F2M (NIPS 2021)  \cite{F2M}  & 72.05 & 67.47 & 63.16 & 59.70 & 56.71 & 53.77 & 51.11 & 49.21 & 47.84 & 57.89 & +27.18\\
    MetaFSCIL (CVPR 2022)  \cite{MetaFSCIL} & 72.04 & 67.94 & 63.77 & 60.29 & 57.58 & 55.16 & 52.90 & 50.79 & 49.19 & 58.85 & +26.22\\
    C-FSCIL (CVPR 2022)  \cite{C-FECIL} & 76.40 & 71.14 & 66.46 & 63.29 & 60.42 & 57.46 & 54.78 & 53.11 & 51.41 & 61.61 & +23.46\\
    FeSSSS (CVPR 2022)  \cite{FeSSSS} & 81.50 & 77.04 & 72.92 & 69.56 & 67.27 & 64.34 & 62.07 & 60.55 & 58.87 & 68.23 & +16.84\\
    ALICE (ECCV 2022)  \cite{ALICE} &80.60 & 70.60 & 67.40 & 64.50 & 62.50 & 60.00 & 57.80 & 56.80 & 55.70 & 63.99 & +21.08\\
    Liu et al. (ECCV 2022) \cite{liu2022few} & 71.84 & 67.12& 63.21 & 59.77 & 57.01 & 53.95 & 51.55 & 49.52 & 48.21 & 58.02 & +27.05 \\
    SReg (ICLR 2022)  \cite{ICLR2022} & 80.37 & 73.76 & 68.36 & 64.07 & 60.36 & 56.27 & 53.10 & 50.45 & 47.55 & 61.59 & +23.48 \\
    LIMIT (TPAMI 2022)  \cite{LIMIT} & 72.32 & 68.47 & 64.30 & 60.78 & 57.95 & 55.07 & 52.70 & 50.72 & 49.19 & 59.06 & +26.01\\
    MCNet (TIP 2023) \cite{ji2023memorizing} & 72.33 & 67.70 & 63.50 & 60.34 & 57.59 & 54.70 & 52.13 & 50.41 & 49.08 & 58.64 & +26.43\\
    NC-FSCIL (ICLR 2023) \cite{yang2023neural} & 84.02 & 76.80 & 72.00 & 67.83 & 66.35 & 64.04 & 61.46 & 59.54 & 58.31 & 67.82 &+17.25\\
    SoftNet (ICLR 2023) \cite{kang2022soft} & 79.77 & 75.08 & 70.59 & 66.93 & 64.00 & 61.00 & 57.81 & 55.81 & 54.68 & 65.07 & +20.00 \\
    SSFE-Net\cite{pan2023ssfe} (WACV 2023) &  72.06 &  66.17 &  62.25 &  59.74 &  56.36 &  53.85 &  51.96 &  49.55 &  47.73 & 57.74 & +27.33 \\
    \hline
    
    \textbf{CEC+V-Swin-T}  & \textbf{88.73} & \textbf{86.74} & \textbf{83.96} & \textbf{82.56} & \textbf{81.46}  & \textbf{79.95}  & \textbf{77.92} & \textbf{78.05} & \textbf{77.52} & \textbf{81.88} &  \textbf{+3.19}\\
    \textbf{CEC+SV-Swin-T}  & \textbf{88.75} & \textbf{87.92} & \textbf{86.07} & \textbf{84.84} & \textbf{84.30} & \textbf{83.24} & \textbf{82.22} & \textbf{82.28} & \textbf{82.38} & \textbf{84.67} & \textbf{+0.40} \\

    \hline

    \textbf{LIMIT+V-Swin-T} & \textbf{89.17} & \textbf{87.39} & \textbf{84.83} & \textbf{83.41} & \textbf{82.66} & \textbf{81.20} & \textbf{79.81} & \textbf{79.36} & \textbf{79.23} & \textbf{83.01} & \textbf{+2.06} \\
    
    \textbf{LIMIT+SV-Swin-T} & \textbf{90.55} & \textbf{89.20} & \textbf{86.80} & \textbf{85.44} & \textbf{84.78} & \textbf{83.38} & \textbf{81.91} & \textbf{81.90} & \textbf{81.65} & \textbf{85.07}  & \\

    \hline
    \end{tabular}
    }
    \label{table:compare FP}
\end{table*}

\begin{figure*}[t]
\begin{minipage}[b]{0.33\linewidth}
  \centering
  \includegraphics[width=\linewidth]{t-SNE-CEC-8.pdf}
  \centerline{(a) CEC}\medskip
\end{minipage}
\begin{minipage}[b]{0.33\linewidth}
  \centering
  \includegraphics[width=\linewidth]{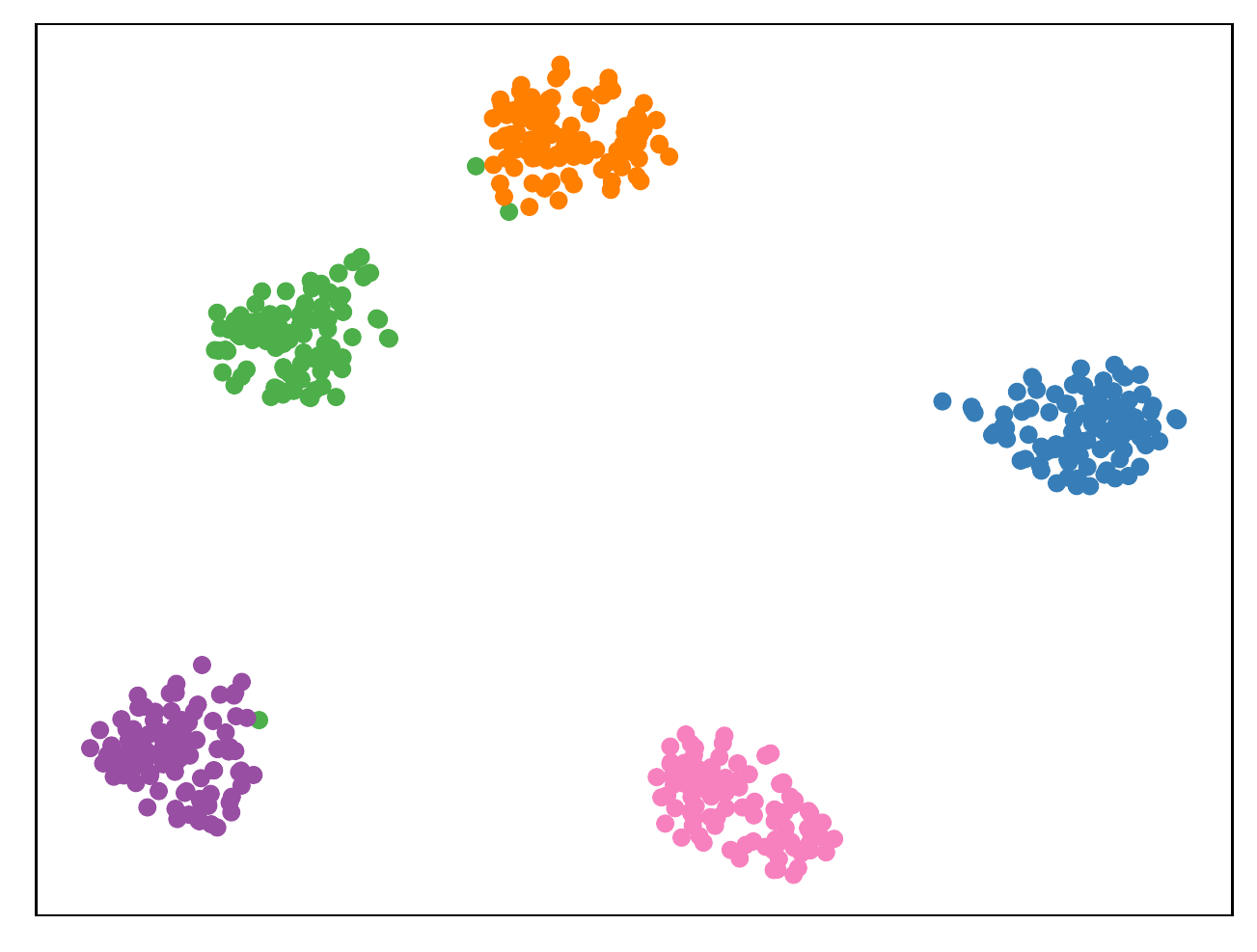}
  \centerline{(b) CEC+SV-ViT}\medskip
\end{minipage}
\begin{minipage}[b]{0.33\linewidth}
  \centering
  \includegraphics[width=\linewidth]{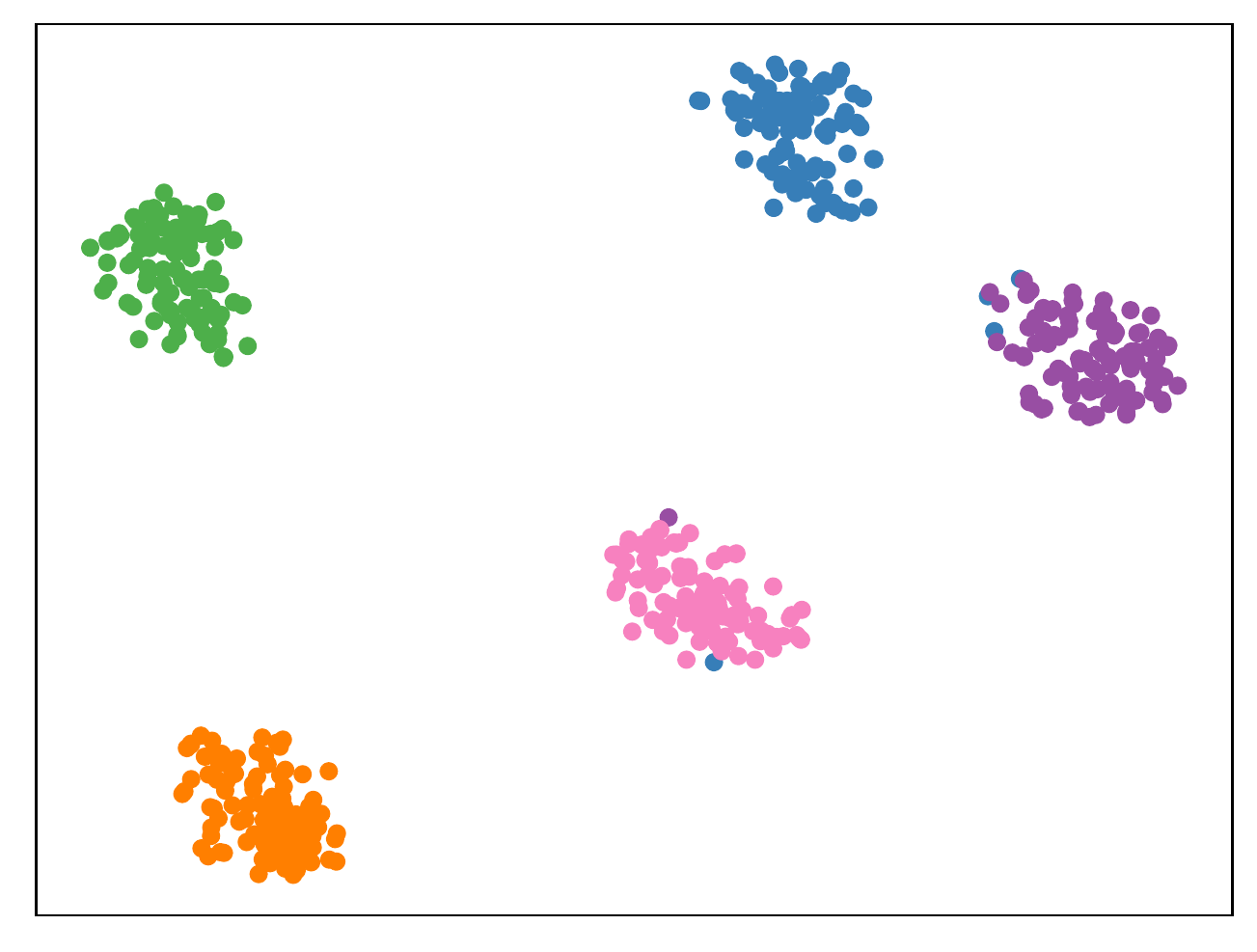}
  \centerline{(c) CEC+SV-Swin-T}\medskip
\end{minipage}

\begin{minipage}[b]{0.33\linewidth}
  \centering
  \includegraphics[width=\linewidth]{t-SNE-LIMIT-8.pdf}
  \centerline{(d) LIMIT}\medskip
\end{minipage}
\begin{minipage}[b]{0.33\linewidth}
  \centering
  \includegraphics[width=\linewidth]{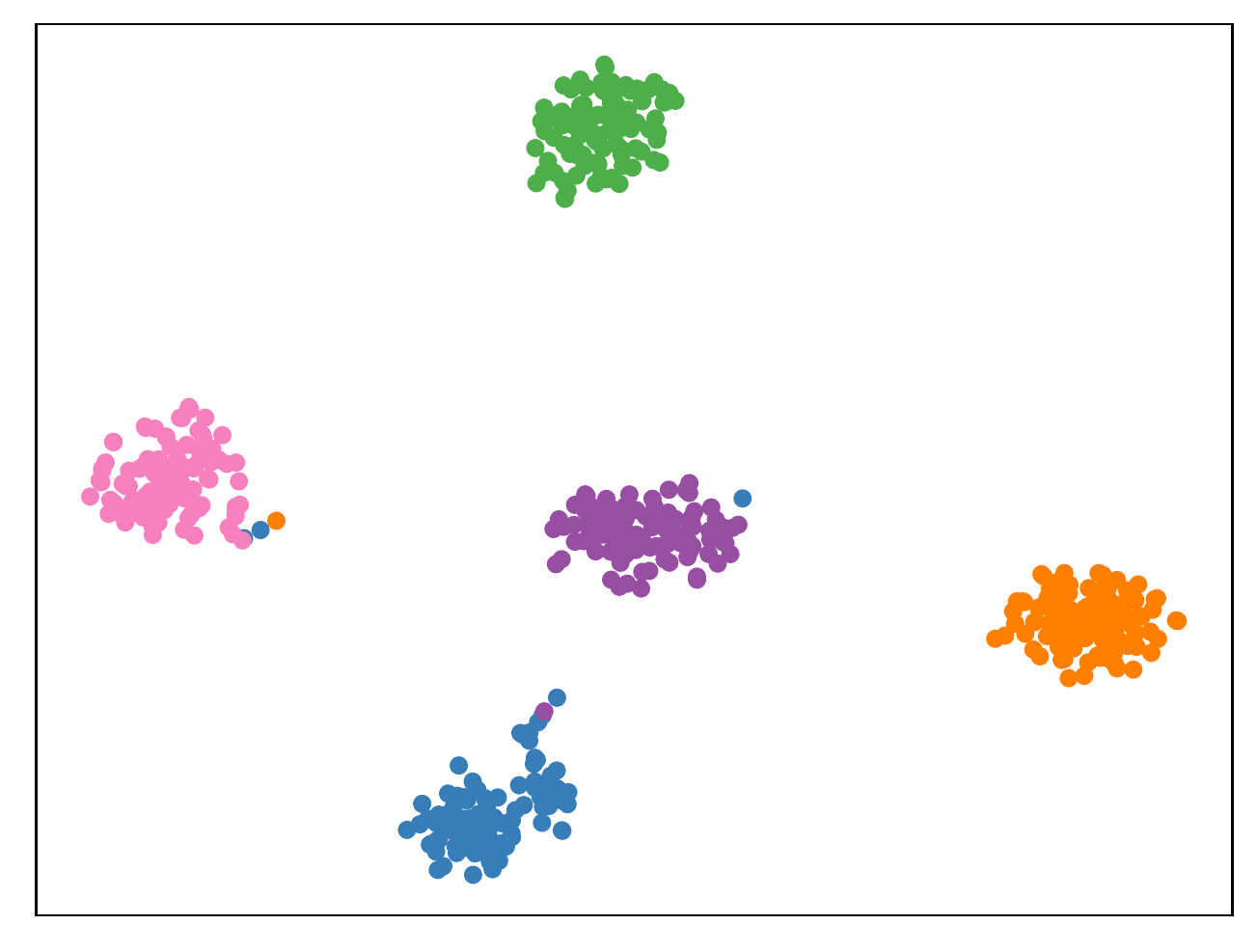}
  \centerline{(e) LIMIT+SV-ViT}\medskip
\end{minipage}
\begin{minipage}[b]{0.33\linewidth}
  \centering
  \includegraphics[width=\linewidth]{t-SNE-Swin-SVT-8.pdf}
  \centerline{(f) LIMIT+SV-Swin-T}\medskip
\end{minipage}
\caption{t-SNE feature embeddings of the 8-th session on the CIFAR100 dataset. Each of the five colors represents a different class: \emph{blue} = whale, \emph{green} = willow tree, \emph{purple} = wolf, \emph{orange} = woman and \emph{pink} = worm.}
\label{img}
\end{figure*}

\bibliographystyle{IEEEtran}
\bibliography{arXiv}

\end{document}